\documentclass{article}

\usepackage{arxiv}

\usepackage[utf8]{inputenc} 
\usepackage[T1]{fontenc}    
\usepackage{hyperref}       
\usepackage{url}            
\usepackage{booktabs}       
\usepackage{amsfonts}       
\usepackage{nicefrac}       
\usepackage{microtype}      
\usepackage{lipsum}		
\usepackage{graphicx}
\usepackage{natbib}
\usepackage{doi}
\usepackage{xcolor,colortbl}
\usepackage{subfigure}
\usepackage{amsmath}

\title{DeepTeeth: A Teeth-photo Based Human Authentication System for Mobile and Hand-held Devices}

\author{Geetika Arora, Rohit K Bharadwaj, and Kamlesh Tiwari\\ Birla Institute of Technology and Science Pilani, Pilani, Jhunjhunu, Rajasthan, 333031 INDIA\\{\tt\small \{p2016406, f20170633, kamlesh.tiwari\}@pilani.bits-pilani.ac.in}}

\renewcommand{\shorttitle}{\textit{arXiv} Template}

\begin{document}
\maketitle

\begin{abstract}
	This paper proposes teeth-photo, a new biometric modality for human authentication on mobile and hand held devices. Biometrics samples are acquired using the camera mounted on mobile device with the help of a mobile application having specific markers to register the teeth area. Region of interest (RoI) is then extracted using the markers and the obtained sample is enhanced using contrast limited adaptive histogram equalization (CLAHE) for better visual clarity. We propose a deep learning architecture and novel regularization scheme to obtain highly discriminative embedding for small size RoI. Proposed custom loss function was able to achieve perfect classification for the tiny RoI of $75\times 75$ size. The model is end-to-end and  few-shot and therefore is very efficient in terms of time and energy requirements. The system can be used in many ways including device unlocking and secure authentication. To the best of our understanding, this is the first work on teeth-photo based authentication for mobile device. Experiments have been conducted on an in-house teeth-photo database collected using our application. The database is made publicly available. Results have shown that the proposed system has perfect accuracy.
\end{abstract}


\section{Introduction}

    Dental biometrics is popularly used to identify victims of natural disasters such as earthquake, cyclone $etc$. Dental structures being one of the hardest parts in the body are able to resist decomposition and extreme temperatures and therefore are among the last to degrade after death \cite{krishan2015dental}. These characteristics of dental structures make them ideal for the application in forensic dentistry. Special equipment and settings are required to acquire dental samples and it becomes difficult to be used for regular authentication process. This work proposes a simple setting of using teeth photo for the authentication. It has been found that the system has higher accuracy as compared to other existing modalities like touch less fingerprint and face. The system also fights against spoof attacks such as \cite{stephanie2002spoofing} for fingerprint and \cite{erdogmus2014spoofing} for face recognition.   
   
           \begin{figure}[t]
            \begin{center}
                \subfigure[Acquisition]
                {\label{fig:teethAcquisitionA}  \includegraphics[width = 3.6cm, height = 4.3cm]{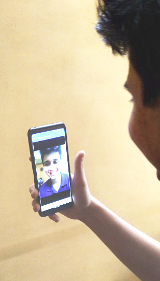}}
                \subfigure[APP interface]
                {\label{fig:teethAcquisitionB}
                    \includegraphics[width=2.9cm,height=4.3cm]{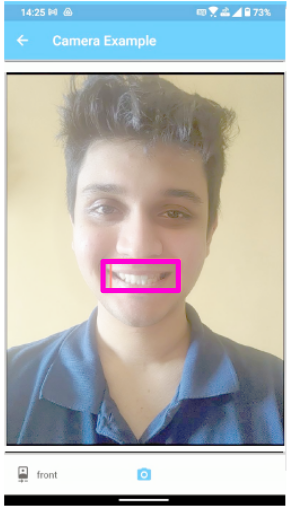}}
                \subfigure[Teeth RoI]
                {\label{fig:teethAcquisitionC}
                    \includegraphics[width=4.3cm,height=4.3cm]{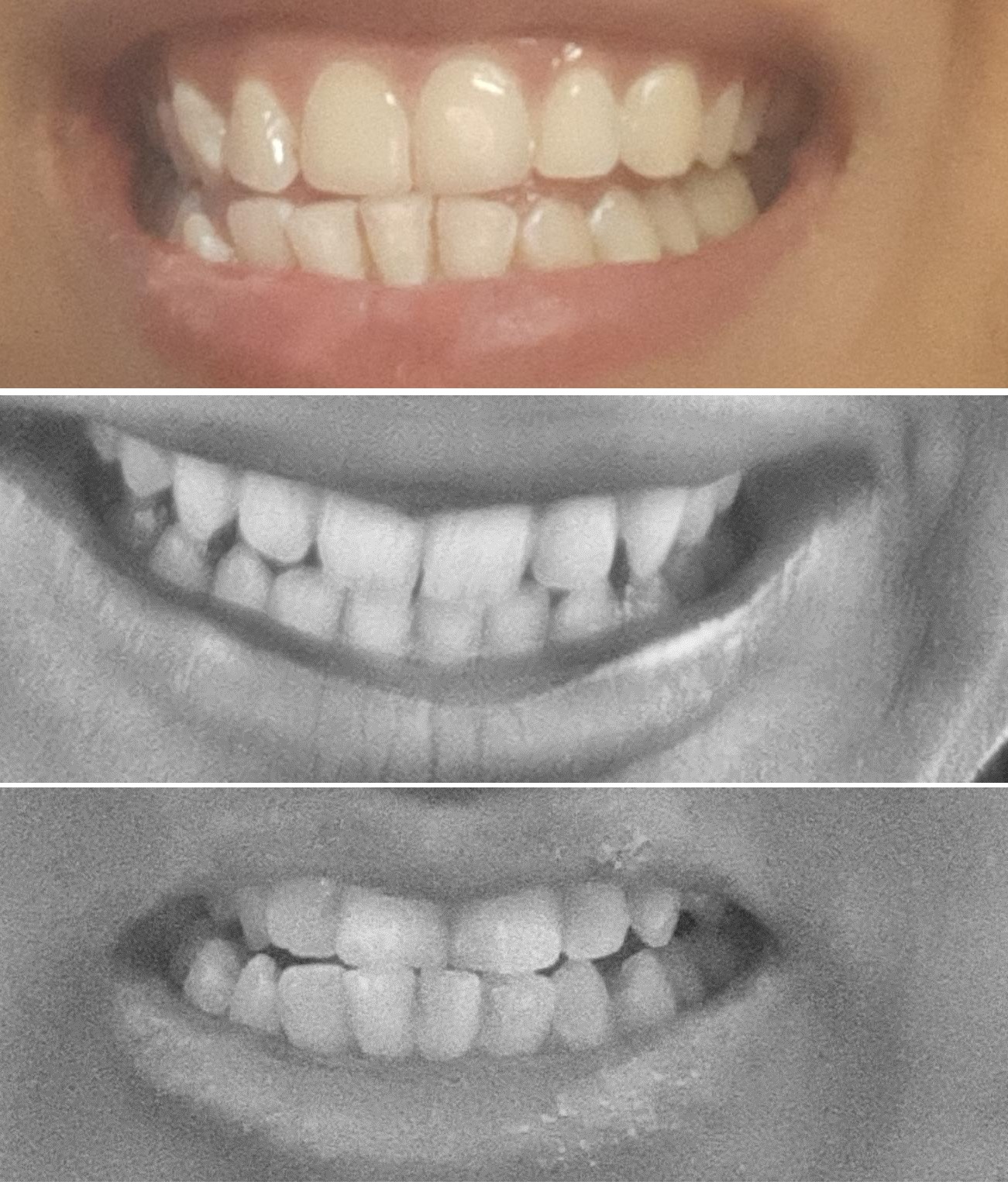}}
            \end{center}
            \caption{Proposed teeth-photo acquisition and region of interest.} \label{fig:teethAcquisition}
        \end{figure}  
        
    Many popular biometric authentication systems in literature are commonly based on traits such as face \cite{givens2013biometric}, fingerprint \cite{jain2016fingerprint}, iris \cite{latman2013field}, voice \cite{feng2017continuous}, retina \cite{sadikoglu2016biometric} $etc$. Dental biometrics has introduced by \cite{jain2003dental,chen2005dental} by using dental radiographs. The purpose of the same was towards forensics and was using dental impressions rather than the direct photo of the same. Jaw and teeth segmentation on X-Ray images for human authentication is done in \cite{bozkurt2020jaw}. Specific oral and maxillo facial identifiers are used in \cite{du2021specific}. A teeth based recognition was used in \cite{kim2008multimodal}, which uses individual identification where they combined teeth and voice features to obtain authentication system with EER of $2.13\%$. All these methods are invasive and suitable for forensic research to identify human remains in accidents and crime sites \cite{pittayapat2012forensic}. They are by design not suitable for regular biometric authentication. More recently, work by \cite{jiang2020smileauth} employ dental features for user authentication. They have obtained the features through first utilizing landmark detection \cite{king2009dlib} and then employing traditional machine learning based methods like Random Forest \cite{ho1995random} for feature extraction. We present teeth-photo based human authentication that works on photos of the teeth taken directly from the dace. For being suitable for mobile devices, there is a need that such a system be efficient in cropping RoI and fast in feature extraction. Such a system should work with small size camera images for better adaptability and power saving. 
        
    \noindent \textbf{Contribution.} 1) We propose teeth-photo as biometric modality that takes images from mounted camera of mobile device instead of radio-graphs or X-ray thereby being more convenient to the user. 2) Devices an application for sample acquisition having markers for efficient RoI extraction. 3) Proposes a deep learning architecture and novel loss function that regularize the latent space such that tiny size images produce perfect results. 4) Collected a teeth-photo database that would be made public. 5) Compared results with other standard deep learning and handcrafted approaches that are scale \& rotation invariant. The results show that the proposed approach gets best results.
  
    Rest of the paper is organized as follows: Next section discusses the proposed approach in detail. Section~\ref{sec:expResults} discusses the results obtained along with a description of the experimental setting, considered datasets and evaluation parameters. Conclusions are presented at the end.
    
    \section{Proposed Approach} 
\begin{figure}
            \centering
            \includegraphics[scale=1]{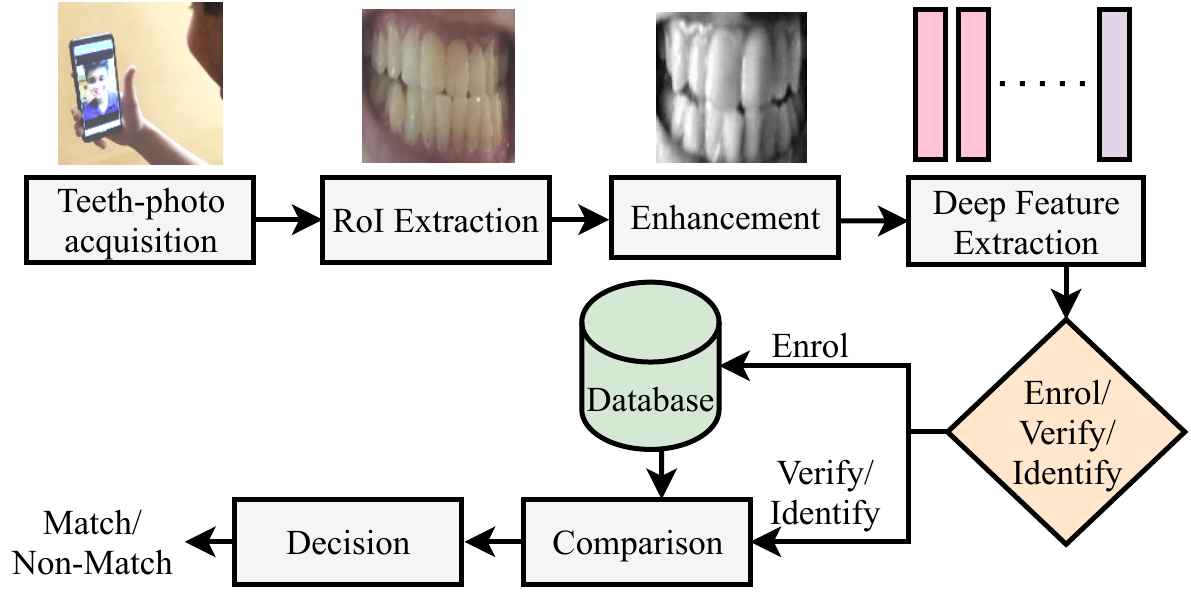}
            \caption{Block diagram of the proposed DeepTeeth approach}
            \label{fig:blockDiag}
        \end{figure}

     This section explains the proposed approach in detail. The approach has six important components 1) sample acquisition, 2) RoI extraction, 3) Enhancement, 4) Deep feature extraction, 5) Comparison, and 6) Decision.
     It starts by sample acquisition capturing photo of teeth of the subject using a mobile application that shows markers on the screen. Refer the \figurename~\ref{fig:teethAcquisitionA} that shows how front-view camera feed is used to get teeth image. On-screen markers for the better acquisition of teeth structure is shown in the \figurename~\ref{fig:teethAcquisitionB}. Region of Interest (RoI) is segmented out from the acquired images and is enhanced using CLAHE. The acquired images are then passed to the proposed network for feature extraction. The extracted features are further used for comparison and decision. Block diagram depicting these steps is shown in \figurename~\ref{fig:blockDiag}.
     
    \begin{figure}[b]
            \begin{center}  
                \includegraphics[scale=0.19]{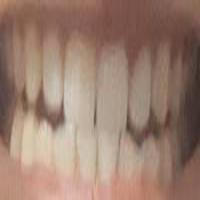}
                \includegraphics[scale=0.19]{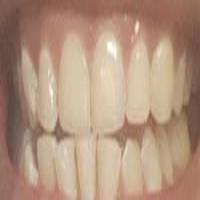}
                \includegraphics[scale=0.19]{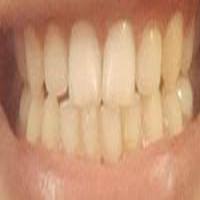}
                \includegraphics[scale=0.19]{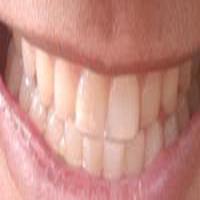}
                \includegraphics[scale=0.19]{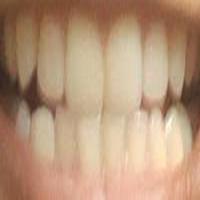}
                \includegraphics[scale=0.19]{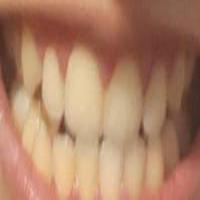}
                \includegraphics[scale=0.19]{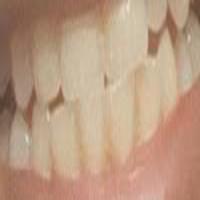}
                \includegraphics[scale=0.19]{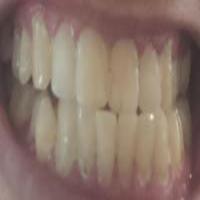}
                \includegraphics[scale=0.19]{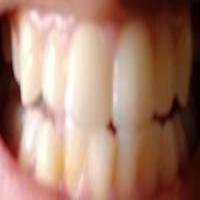}
                \includegraphics[scale=0.19]{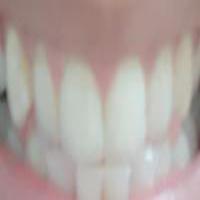}
                \includegraphics[scale=0.19]{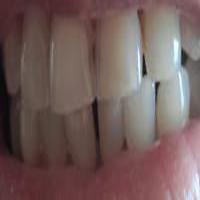}

                \includegraphics[scale=0.19]{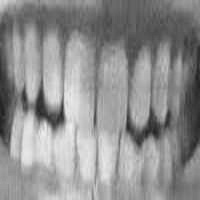}
                \includegraphics[scale=0.19]{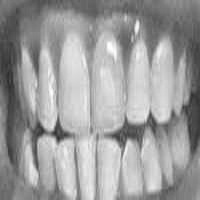}
                \includegraphics[scale=0.19]{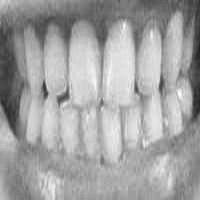}
                \includegraphics[scale=0.19]{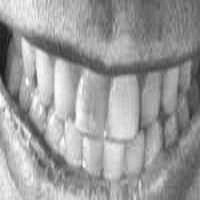}
                \includegraphics[scale=0.19]{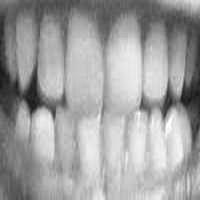}
                \includegraphics[scale=0.19]{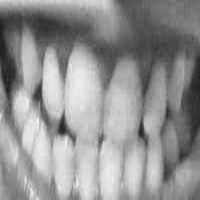}
                \includegraphics[scale=0.19]{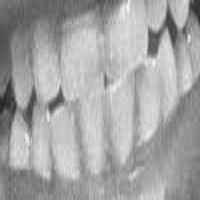}
                \includegraphics[scale=0.19]{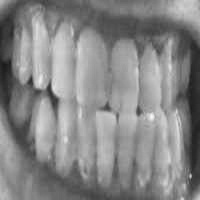}
                \includegraphics[scale=0.19]{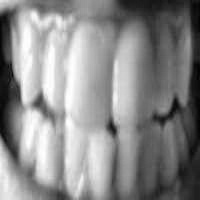}
                \includegraphics[scale=0.19]{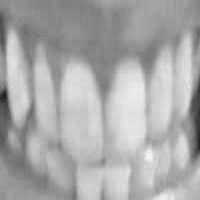}
                \includegraphics[scale=0.19]{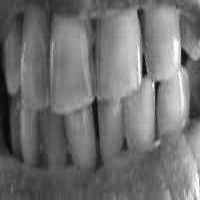}
            \end{center}
            \caption{Normalized RoI images and their corresponding enhancement shown in first and second row respectively. \label{fig:dbImgs}}
        \end{figure}

    \noindent\textbf{Teeth-photo Acquisition.}
       The system captures photos using the camera available on the mobile device. A specially designed android application is used that provided a rectangular marker to guide the user to align their teeth in the right way. These markers also helps in effectively segmenting the region of teeth later. \figurename~\ref{fig:teethAcquisitionA} shows a teeth photo acquisition setting where as \figurename~\ref{fig:teethAcquisitionB} shows the corresponding android application interface.
      
      \noindent\textbf{RoI Extraction.}
        Teeth-photo images of the subjects are taken using the front-view camera of the mobile device. The user holding mobile device have a lot of freedom in terms of scale and rotation. Acquired frame could have background and different lightening conditions. However the rectangular marker on the android application solves the localization problem of the teeth area. The rectangle is $.375\times (width)$ from the left and   $0.5\times (height)$ from the top of the screen. The width and height of the rectangle are  $.5\times (width)$ and $0.12\times (height)$ respectively. Acquired images are cropped accordingly using this rectangle frame coordinate to obtain a  $1416\times 510$ size image as shown in  the \figurename~\ref{fig:teethAcquisitionC}. The rectangular image is further resized to get square size RoI using bi-cubic interpolation. Some of the sample RoI images are shown in \figurename~\ref{fig:dbImgs}.

    \noindent\textbf{Enhancement.}
       RoI image is subjected to gray scale conversion than then contrast of the same is enhanced by applying histogram equalization. We have used contrast limited adaptive histogram equalization\cite{zuiderveld1994contrast} which improves the local contrast and enhances the edge definitions in every region of the image. This contrast amplification is limited to reduce the noise amplification. In the process, RoI is divided into small blocks called tiles. We have considered $8\times8$ tiles. Histogram equalization is done in each of these tiles separately. If any histogram bin is above the specified contrast limit, which in case was set at $2.0$ those pixels are clipped and uniformly distributed to other bins before applying CLAHE.  This enhancement compensates for the dim lighting or reflection if any while taking the teeth photos. The enhanced image is therefore more suitable for the feature extraction and matching. Few sample RoI images and their enhanced versions are  shown in \figurename~\ref{fig:dbImgs}.

    \subsection{Deep Feature Extraction}
    
       Softmax \cite{liu2016large} is the most commonly used loss function for training a classification network to obtain high intra-class similarity. Consider $N$ training examples $X=\{ x_1, x_2, ..., x_N\}$ and their corresponding features $L = \{l_1, l_2, ..., l_N \}$ with $C$ number of classes. The softmax loss is average of negative log probabilities as  defined in Equation~\ref{eq:softMax}, where $y_i$ is the predicted output for $x_i$. 
      
      \begin{equation} \label{eq:softMax}
          L_{sm} = \frac{1}{N} \sum_{i=1}^{N} \left ( -\log \frac{e^{f_{y_i}}}{\sum_{j=1}^{C} e^{f_{j}}}  \right )
      \end{equation}
      
      \noindent Here $f_j$ represents activation of the softmax layer corresponding to $j^{th}$ class. The value of $f_j$ is the product of weight matrix with the feature $f_{j} = W_{j}^{T} \cdot l$. Since any scalar multiplication with the weight vector does not affect the decision boundary we may safely fix $\left\|W_j\right\|$ equal to unity to facilitate effective learning. This leads us to $f_{j} = \left\|l\right\|\cos\theta_j$, where $\theta_j$ is the angle between $W_j$ and $l$. So the same loss function could be re-written as Equation~\ref{eq:softMax2}
      
    \begin{equation}\label{eq:softMax2}
         L_{sm} = \frac{1}{N} \sum_{i=1}^{N} \left ( -\log \frac{e^{\left \| l_i \right \| \; \cos\theta_{y_i}}}{\sum_{j=1}^{C} e^{\left \| l_j \right \| \; cos\theta_{j}}}  \right )
    \end{equation}
      
      \begin{figure}
            \centering
            \includegraphics[scale=0.8]{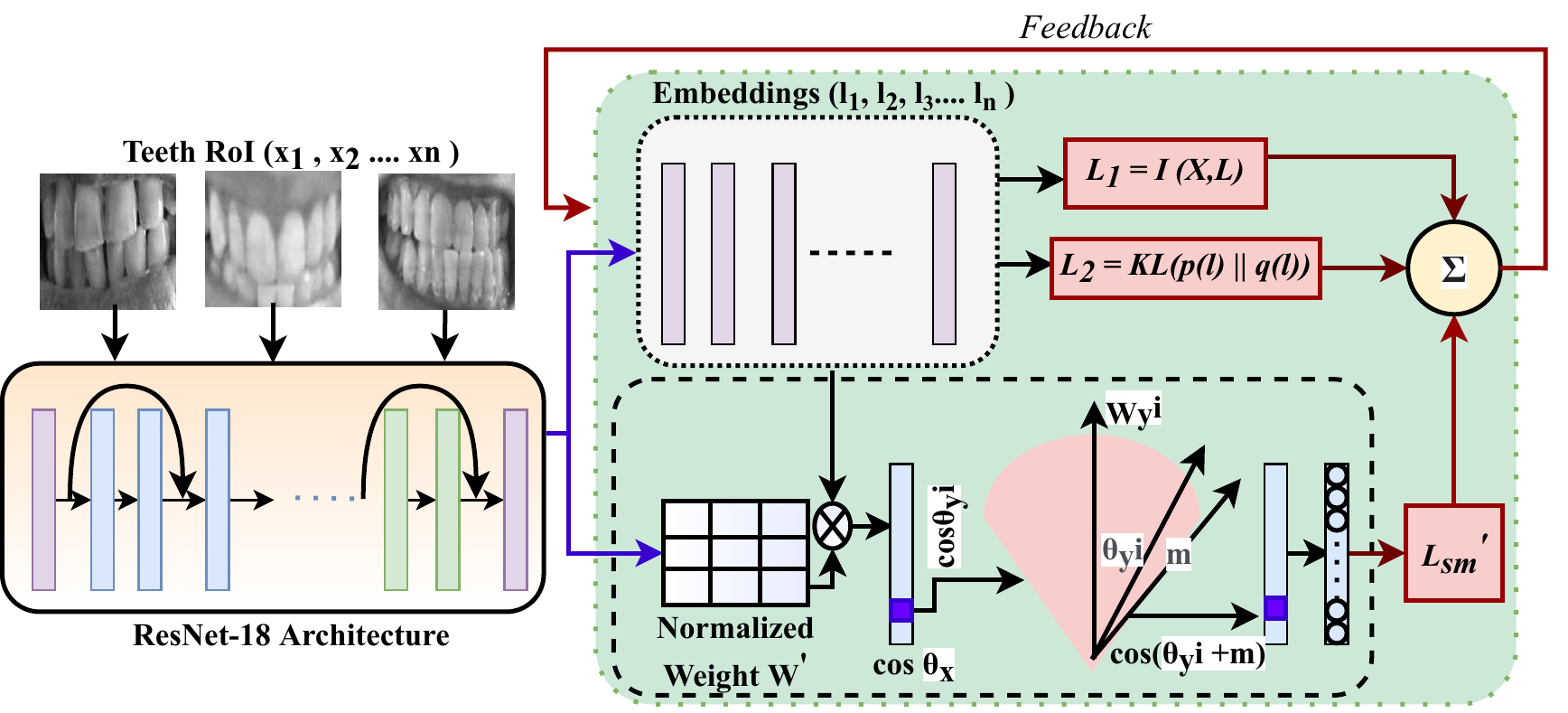}
            \caption{Proposed Teeth-photo architecture and loss function.}
            \label{fig:proposedArchLoss}
        \end{figure}
    
    This equation have focus on angular disparity therefore, feature embedding $l$ could also be scaled to a fixed value $\|l\|=s$ like the weight matrix $W$ without loosing any discrimination ability. Further to improve the separation between latent representations of the different classes we add a margin of $m$ to the angular similarity in the $\cos\theta$ term \cite{deng2019arcface}. So the modified softmax loss function is as  per the Equation~\ref{eq:softMax3}. 
     
    \begin{equation} \label{eq:softMax3}
          {L}'_{sm} =\frac{-1}{N} \sum_{i=1}^{N} \log \frac{e^{s \cos\left ( \theta_{y_i} + m  \right )}}{e^{s\cos\left (\theta_{y_i} + m  \right )} + \sum_{j=1, !y_i}^{C} e^{s \cos\theta_{j}}} 
    \end{equation}
     
    This arrangement works against blurred boundaries between the classes and enhance the inter-class distance. This helps to handle pose, illumination and background variations. Correlation between the learned latent representation and its corresponding input sample could be established by maximizing the mutual information between the two.   Mutual information between samples $x$ and its feature $l$ can be seen as the probability of obtaining the value $l$ given $x$. We wish its sum over all the data points to be maximized. As stated in Equation~\ref{eq:eq2} mutual information can be estimated using KL-divergence \cite{goldberger2003efficient}.  
        
\begin{equation} \label{eq:eq2}
\begin{aligned}
I(x, l) = \int\int p(l|x)p(x) \log\frac{p(l|x)}{p(l)} dx dl = KL(p(l|x) p(x) || p(l) p(x))
\end{aligned}
\end{equation}
      
      Also, to make the embedding space more regular, the distribution of the latent representations, denoted by $p(l)$ should follow a prior distribution such as that of the normal distribution, denoted by $q(l)$. This is achieved by minimizing the KL divergence between $\phi(l)$ and $q(l)$ (Equation~\ref{eq:eq3}).
        
        \begin{equation} \label{eq:eq3}
           KL(p(l) || q(l)) = \int p(l) \log\frac{p(l)}{q(l)}\; dl
        \end{equation}
        
      We need to maximize the mutual information or the correlation between input sample and its corresponding learned representation along with minimizing the KL divergence between the the prior distribution and distribution followed by the latent representations. Hence, combining Equation~\ref{eq:eq2} and Equation~\ref{eq:eq3},
        
        \begin{equation} \label{eq:eq4}
          p(l|x) = min \{- I(X,L) + \mathbb{E}_{x \sim p(x)}[ KL(p(l) || q(l))]\}
        \end{equation}
        
        The KL divergence can take any value between zero to infinity with zero meaning that the two distributions are similar. Issue that the KL divergence not having a well defined upper bound \cite{rai2018visual} could be handled using JS divergence \cite{wang2006groupwise} instead. This changes the Equation~\ref{eq:eq4} to Equation~\ref{eq:eq5}.

   \begin{table}[]
       \centering
       \caption{Performance of the system on different size of teeth image RoI. ERR and EER are shown as percentage (\%). Number of falsely accepted (FA) and falsely rejected (FR) cases are reported out of 204 and 10200 comparisons for 256 dimensional fixed size feature vector of deep learning approaches.  Higher CRR and low EER is better. }
       \label{tab:my_labelTableCompareAll}
       \begin{tabular}{lcccccc}  \hline
       \rowcolor{gray!20!white}
       & \textbf{Size}& \textbf{CRR} & \textbf{EER} &\textbf{DI} & \textbf{FR} & \textbf{FA} \\\hline\hline
            ORB & 25$\times$25 & 11.76 & 41.966 & 0.30 & 87 & 4211 \\\hline
            ORB & 50$\times$50 & 67.65 & 22.637 & 1.08 & 46 & 2318 \\\hline
            ORB & 75$\times$75 & 71.57 & 18.157 & 1.23 & 37 & 1854 \\\hline
            ORB & 100$\times$100 & 66.67 & 22.064 & 1.04 & 45 & 2251 \\\hline
            ORB & 125$\times$125 & 84.31 & 16.181 & 1.32 & 33 & 1651 \\\hline
            ORB & 150$\times$150 & 81.37 & 14.706 & 1.39 & 30 & 1500 \\\hline
            ORB & 175$\times$175 & 81.37 & 16.672 & 1.32 & 34 & 1701 \\\hline
            
            SIFT & 25$\times$25 & 11.76 & 40.833 & 0.33 & 83 & 4180 \\\hline
            SIFT & 50$\times$50 & 39.22 & 29.044 & 0.80 & 59 & 2975 \\\hline
            SIFT & 75$\times$75 & 41.18 & 23.049 & 0.99 & 47 & 2352 \\\hline
            SIFT & 100$\times$100 & 31.37 & 29.348 & 0.83 & 60 & 2987 \\\hline
            SIFT & 125$\times$125 & 49.02 & 21.946 & 1.08 & 45 & 2227 \\\hline
            SIFT & 150$\times$150 & 57.84 & 21.569 & 1.12 & 44 & 2200 \\\hline
            SIFT & 175$\times$175 & 39.22 & 23.020 & 1.02 & 47 & 2346 \\\hline
            
            Siam & 25$\times$25 & 89.22 & 4.412 & 2.14 & 9 & 450 \\\hline
            Siam & 50$\times$50 & 98.04 & 0.892 & 2.80 & 2 & 82 \\\hline
            Siam & 75$\times$75 & 100.00 & 0.608 & 2.79 & 1 & 74 \\\hline
            Siam & 100$\times$100 & 100.00 & 0.980 & 2.55 & 2 & 100 \\\hline
            Siam & 125$\times$125 & 96.08 & 1.093 & 2.19 & 2 & 123 \\\hline
            Siam & 150$\times$150 & 99.02 & 1.034 & 2.03 & 2 & 111 \\\hline
            Siam & 175$\times$175 & 98.04 & 1.525 & 1.83 & 3 & 161 \\\hline
            
            Ours & 25$\times$25 & 100.00 & 0.000 & 4.89 & 0 & 0 \\\hline
            Ours & 50$\times$50 & 100.00 & 0.088 & 4.18 & 0 & 18 \\\hline
            \rowcolor{green!30}
            Ours & 75$\times$75 &100.00 &0.000 & 3.82 & 0 & 0 \\\hline
            Ours & 100$\times$100 & 100.00 & 1.471 & 3.07 & 3 & 150 \\\hline
            Ours & 125$\times$125 & 100.00 & 2.451 & 3.06 & 5 & 250 \\\hline
            Ours & 150$\times$150 & 100.00 & 1.471 & 2.83 & 3 & 150 \\\hline
            Ours & 175$\times$175 & 100.00 & 2.471 & 2.85 & 5 & 254 \\\hline
            Ours & 200$\times$200 & 99.02 & 3.431 & 2.56 & 7 & 350 \\\hline
     \end{tabular}
   \end{table}

    \begin{eqnarray} \label{eq:eq5}
             p(l|x) &=& \mathop{min} (- JS(p(l|x) p(x), p(l)p(x)) \nonumber\\
              && + \mathbb{E}_{x \sim \phi(x)}[ KL(p(l) || q(l))])
    \end{eqnarray}

    The total loss of the proposed network is computed by combining  Equation~\ref{eq:softMax3} and Equation~\ref{eq:eq5}.
    
    \begin{eqnarray} \label{eq:14}
        L  & =& -\alpha (JS(p(l|x) p(x), p(l)p(x))) \nonumber\\
              && + \mathbb{E}_{x \sim \phi(x)}[ KL(p(l) || q(l))]  \nonumber \\
              && -\frac{\gamma}{N} \sum_{i=1}^{N} \log \frac{e^{s \cos\left ( \theta_{y_i} + m  \right )}}{e^{s\cos\left (\theta_{y_i} + m  \right )} + \sum_{j=1, !y_i}^{C} e^{s \cos\theta_{j}}} 
    \end{eqnarray}

    For the set of biometric samples $X = \{x_1, x_2, ..., x_n\}$ that are passed to a classification network; learned latent representation $L= \{l_1, l_2, ..., l_n\}$ are obtained from the second last layer, which is before the softmax layer. The diagram in  \figurename~\ref{fig:proposedArchLoss} shows the proposed loss and feature construction.
        
        \begin{figure}
            \centering
            \includegraphics[scale=0.21]{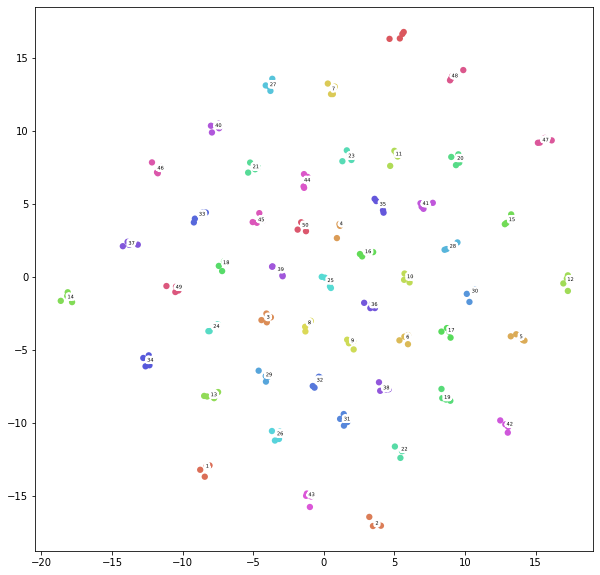}
            \includegraphics[scale=0.35]{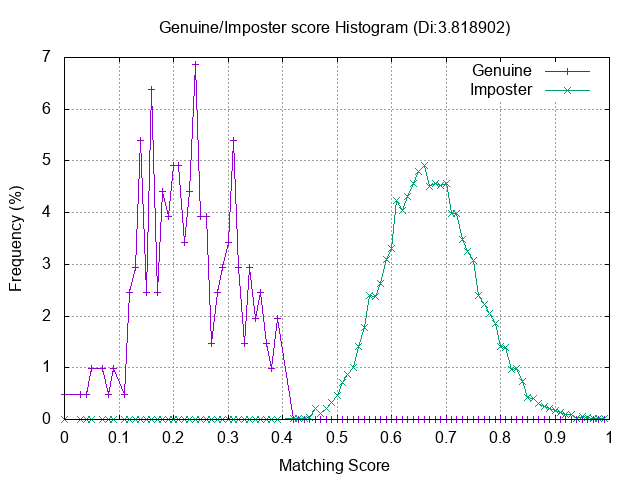}
            \includegraphics[scale=0.35]{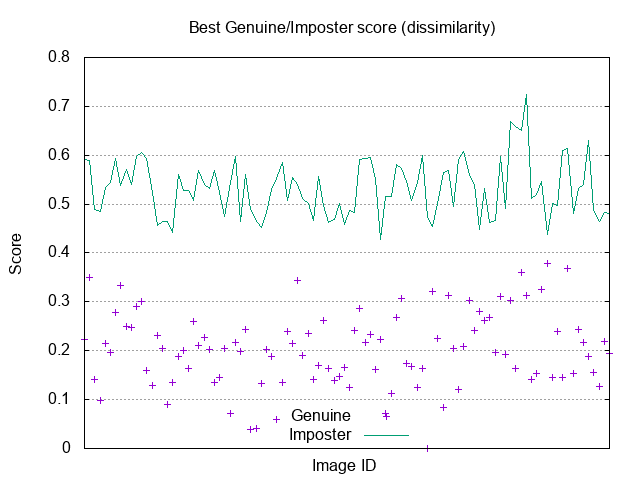}
            \caption{Plots showing, 1) t-SNE plot, 2) Imposter/Genuine score histogram, and 3) best genuine and imposter scores of all the images for $75\times 75$ size normalized RoI.}
            \label{fig:my_labelROIANdOthePlot}
        \end{figure}
        
    \noindent\textbf{Comparison.}
      The teeth recognition system takes a query teeth image as an input and compares it with the template stored against the claimed identity. Distinguishing features embedding are obtained using the proposed network for the claimed and query sample and Euclidean distance is obtained. The score serves as a dissimilarity measure for the analysis.  
        
    \noindent\textbf{Decision.} 
      Decision to determine whether the two samples are of the same person or not is always based on the threshold that is empirically determined. If the system is to be deployed in high security area then the threshold could he hardened by selecting lower value for dissimilarity. For low security places the threshold could be relaxed by allowing higher dissimilarity value. However, a fair idea could be obtained of the system performance if we choose the threshold that is at equal err or rate (ERR) that refers the point on ROC curve where true positive rate and true negative rates are equal. 
 
\section{Results \label{sec:expResults}}  
   This section explains the experimental setup, parameters used, and the database used in the experiment. Observations of results obtained, ablation study and discussions are presented. 
   
   \subsection{Database}
    An in-house database of the teeth-photo has been collected from 51 users. This much number of users is moderate for number of persons using single handheld device. A cross platform application was specially developed on \texttt{Android} operating system for the data collection. Application had a rectangular marker for the proper positioning of the teeth in the frame as shown in \figurename~\ref{fig:teethAcquisitionB}. The mobile device had 6GB RAM and was running Android version 9 with a camera with resolution of 16MP.  Every user has provided four sample teeth images within a span of 3-4 days. Some of the database images are shown in \figurename~\ref{fig:dbImgs}
   
   \noindent\textbf{The Siamese Network.}
     The Siamese network is trained for comparison with proposed method using triplet loss \cite{DBLP:journals/corr/SchroffKP15}. The model is set to reduce euclidean distance between the embedding of positive class and increase for negative class. The two similar networks of siamese network contains pair of convolution layers with varying filter of size $128$ with $7\times7$ and $3\times3$. The initialization of weights have been adopted, and $L_2$ weight regularization of $2\times10^{-4}$ and $1\times10^{-3}$ have been used for different layers of the network.
     Adam optimizer with learning rate of $6\times10^{-5}$, $\beta1$ value 0.9, $\beta2$ value 0.999 and $\epsilon=0.1$ has been used. The minimum margin $\alpha$ was set at 0.2. Batch size for the images has been taken as 16 and trained for 250 epochs. From a large batch of 2500 Triplets, 150 hard and 200 random triplets were selected to train Siamese Network
        
   \begin{figure}[t]
            \begin{center}  
                \includegraphics[scale=0.5]{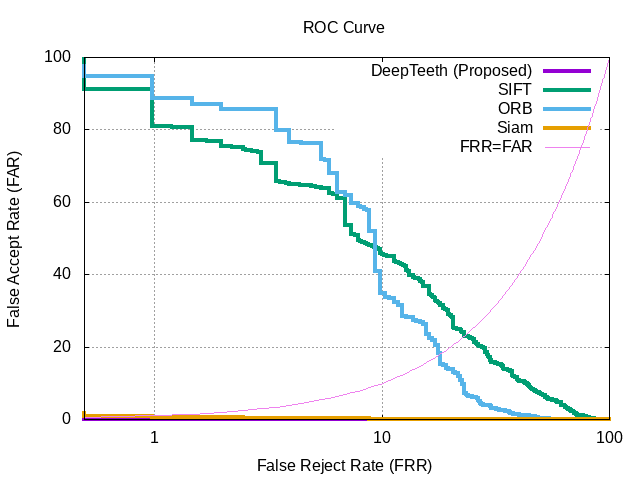}
            \end{center}
            \caption{ROC curve of SIFT \cite{lowe2004distinctive}, ORB \cite{rublee2011orb}, Siamese \cite{koch2015siamese} and the proposed system for $75\times 75$ size normalized RoI. FAR is at $\log$ scale} \label{fig:ROC}
        \end{figure}
   
   \noindent \textbf{Experimental Setting.}
     We normalize and randomize the dataset for creating the input batch. The backbone network ResNet-18 \cite{he2016identity} is utilized to get 256 dimensional embedding vector for each teeth image. These embedding were then optimized using $s = 30$, and $m = 0.35$. During testing, two teeth feature vector are compared using Euclidean similarity. The value of $\gamma$ is set to $2$. Batch size of $16$ was taken and the proposed methodology was trained for $100$ epochs using Adam optimizer and $L_2$ weight regularization of $5\times10^{-4}$. Initially, the learning rate was set at $10^{-3}$ and was decayed every $20$ epochs by $\gamma = 0.1$. The model was trained on Dell Inspiron-15-5577, on Nvidia GTX 1050 GPU the training took around 24.7 minutes. This finally produces a 256-dimensional feature vector.
  
   \noindent \textbf{Performance Measures.} We have used five parameters to compare the system. \emph{Correct Recognition Rate (CRR)} is the ratio of number of subjects correctly classified to the total number of test subjects. 
   \emph{Equal error rate (ERR)} is the point on the ROC curve where proportion of false acceptances is same as the proportion of false rejection. The lower this value, the better the system. Higher CRR and low EER is better. \emph{Error Under Curve (EUC)} is a scalar which gives the probability that a classifier outputs higher match score to a randomly picked genuine sample than to randomly picked imposter sample. \emph{Accuracy} is the maximum value of $(100-\frac{FRR+FAR}{2})$ for all thresholds.  \emph{Discriminative Index} ($DI$) is defined as  $=\frac{|\mu_{Genuine}-\mu_{Imposter}|}{\sqrt{\sigma_{Genuine}^2+\sigma_{Imposter}^2}}$  represents the separation between genuine and imposter scores.

   \subsection{Results and Discussion}
      The proposed approach has obtained best results on the normalized RoI image of size $75\times75$. In this setting total number of comparison scores obtained are 10404  that includes 204 genuine and 10200 imposter scores. In this setting there was no false accept and false reject so the system have achieved CRR=100\% and EER=0.0\%. Mean of the genuine scores is found to be 0.233642 and that of imposter scores is 0.678135. Standard deviation of the matching score is be 0.081486 and 0.083111 for genuine  and  imposter respectively. Discriminative Index ($DI$), which represents the separation between genuine and imposter scores is found to be 3.818. Receiver operating characteristic curve (ROC) of the system is shown in  \figurename~\ref{fig:ROC}. Genuine and Imposter score histogram for the setting are plotted in \figurename~\ref{fig:my_labelROIANdOthePlot}(2). As can be seen from the \tablename~\ref{tab:my_labelTableCompareAll} for RoI  size $75\times75$ other methods such as SIFT \cite{lowe2004distinctive}, ORB \cite{rublee2011orb} and Siamese network \cite{koch2015siamese} have CRR as $41.18$\%, $71.57$\% and $100.0$\% respectively, while EER for SIFT, ORB and Siamese network are found to be $23.049$\%, $18.157$\% and $0.608$\% respectively. We adopted a different testing strategy to calculate scores for the best matches.  In this case, for every training image, only two of its matching scores are used out of the total available scores. First one is the minimum of the imposter matching score and second one is the minimum of all genuine scores.
      
        These scores for all images are plotted in \figurename~\ref{fig:my_labelROIANdOthePlot}(c). Effectiveness of the proposed method could also be seen from the $t-SNE$ plot presented in \figurename~\ref{fig:my_labelROIANdOthePlot}(1).
      
       It can be seen that the embedding of same class are near and are far from different classes. There is a clear margin of separation obtained by this method. The accuracy of the proposed approach has been found 100.00\%.
      
      For comparison of the proposed method, three other methods have been used to extract features from the acquired dataset $viz.$ SIFT, ORB, and Siamese. The comparative analysis has been done using the aforementioned evaluation parameters. It has been reported that the accuracy achieved by the SIFT and ORB is 77.039\% and 84.828\% respectively. While, the accuracy was found to be 99.392\%  on Siamese network, it can be observed that 100\% accuracy with 0.00\% EER has been reported for the proposed approach. The same has been reported in \tablename~\ref{tab:my_label}. 
      
      \subsubsection{Ablation Study.} 
         Various methods for feature extraction have been explored to set-up the best possible system for teeth recognition. Handcrafted methods such as SIFT and ORB along with deep learning networks (Siamese network and DeepTeeth) have been utilized for feature learning. Further, the proposed system has been investigated on different values of 'margin' while training. The best margin value achieved is 0.5. After fixing the value of margin, the size of the images has been varied for learning a n-dimensional feature vector corresponding to each teeth image. The size of feature vector has been set to 256. The learned embeddings of different size are then subjected to feature evaluation process wherein, $n^2$ matching is performed to compute CRR, EER and Accuracy of the system. It has been observed that the system performs best with 256-dimension feature with image size of $75\times75$. The same can be observed from \tablename~\ref{tab:my_labelTableCompareAll}.

       \begin{table}[]
        \begin{center}
        \caption{Comparison of the proposed method DeepTeeth with respect to other techniques on normalized RoI 75$\times$75 size.}
            \begin{tabular}{|l|c|c|c|c|}\hline \rowcolor{gray!30}
                & CRR (\%) & EER (\%) & DI & Accuracy  \\\hline  \hline
                SIFT \cite{lowe2004distinctive} &  41.176  & 23.049  & 0.985 & 77.039 (\%)  \\\hline 
                ORB \cite{rublee2011orb} & 71.569   & 18.156  & 1.234 & 84.828 (\%) \\\hline 
                Siamese \cite{koch2015siamese} &  100.00  & 0.607  & 2.788 & 99.392(\%)  \\\hline  \hline
                \textbf{Proposed} & 100.00   & 0.000  & 3.818 & 100.00 (\%) \\\hline  
            \end{tabular}
        \end{center}
        \label{tab:my_label}
      \end{table}

    \section{Conclusion}
      This paper presents teeth-photo a novel biometric trait for human authentication that is acquired using mounted camera of mobile device with the help of markers on Android applications for easy RoI segmentation. It has been observed that the less explored teeth-photo has very high recognition and identification accuracy with the special feature proposed in the paper. The deep learning based model however take longer time to train initially, but at the time of deployment it is highly efficient for identification or verification. Proposed model works perfectly with small size sample therefore is power efficient and suitable for mobile devices. We have also proposed a novel method for the regularization of the deep learning architecture by combining margin and mutual information in the back bone feature representation. Coupled with CLAHE enhancement the system have achieves perfect classification and identification accuracy for tiny RoI image of $75\times 75$ size. The same RoI with other rotation and scale invariant methods such as SIFT and ORB have achieved EER of 29.34\% and 18.15\%. the value of EER for Siamese network was found to be 0.608\%. However, the proposed system achieves EER of 0.0\% that shows the superiority of the constructed features. Correct recognition rate (CRR) of the proposed method is also found to be 100\%. This establishes suitability of teeth-photo as a biometric trait for human recognition.

\bibliographystyle{unsrtnat}
\bibliography{references}  

\end{document}